\documentclass[10pt,twocolumn,letterpaper]{article} 

\usepackage[dvipsnames]{xcolor}

\usepackage{avss}
\usepackage{times}
\usepackage{epsfig}
\usepackage{graphicx}
\usepackage{amsmath}
\usepackage{amssymb}

\usepackage{balance}
\usepackage{multirow,ctable}
\usepackage{hyperref}

\usepackage[hang,flushmargin]{footmisc}

\avssfinalcopy 

\def\avssPaperID{89} 

\DeclareMathOperator{\stack}{stack}

\def\metodo{SkeleMotion}

\ifavssfinal\fi
\begin{document}

\title{\metodo: A New Representation of Skeleton Joint Sequences Based on Motion Information for 3D Action Recognition}

\author
{
	Carlos Caetano$^{1}$ \and
	Jessica Sena$^{1}$ \and
	François Brémond$^{2}$ \and
	Jefersson A. dos Santos$^{1}$ \qquad William Robson Schwartz$^{1}$\\
	$^{1}$Smart Sense Laboratory, Department of Computer Science\\ Universidade Federal de Minas Gerais, Belo Horizonte, Brazil\\
	$^{2}$INRIA, Sophia Antipolis, France\\
	{\tt\small$\{$carlos.caetano, jessicasena, jefersson, william$\}$@dcc.ufmg.br}\\
	{\tt\small francois.bremond@inria.fr}
}

\maketitle
\begin{abstract}
	Due to the availability of large-scale skeleton datasets, 3D human action recognition has recently called the attention of computer vision community. Many works have focused on encoding skeleton data as skeleton image representations based on spatial structure of the skeleton joints, in which the temporal dynamics of the sequence is encoded as variations in columns and the spatial structure of each frame is represented as rows of a matrix. To further improve such representations, we introduce a novel skeleton image representation to be used as input of Convolutional Neural Networks (CNNs), named \metodo. The proposed approach encodes the temporal dynamics by explicitly computing the magnitude and orientation values of the skeleton joints. Different temporal scales are employed to compute motion values to aggregate more temporal dynamics to the representation making it able to capture long-range joint interactions involved in actions as well as filtering noisy motion values. Experimental results demonstrate the effectiveness of the proposed representation on 3D action recognition outperforming the state-of-the-art on NTU RGB+D~120 dataset.
\end{abstract}

\section{Introduction} \label{sec:intro}

Human action recognition plays an important role in various applications, for instance surveillance systems can be used to detect and prevent abnormal or suspicious actions, health care systems can be used to monitor elderly people on their daily living activities and robot and human interactions.

Over the last decade, significant progress on the action recognition task has been achieved with the design of discriminative representations employed to the image and video domains on RGB data or optical flow. Such information is based on appearance or motion analysis. Due to the development of low-cost RGB-D sensors (e.g., Kinect), it becomes possible to employ depth information as well as human skeleton joints to perform 3D action recognition. Compared to RGB and optical flow, skeleton data has the advantages of being computationally efficient since the data size is smaller. Moreover, skeleton data are robust to illumination changes, robust to background noise and invariant to camera views~\cite{Han:2017}.

Many works for 3D action recognition have focused on designing handcrafted feature descriptors~\cite{Wang:2012, Yang:2012, Zanfir:2013, Gowayyed:2013, Wang:2014, Devanne:2015} to encode skeleton data while adopting Dynamic Time Warping (DTW), Fourier Temporal Pyramid (FTP) or Hidden Markov Model (HMM) to model temporal dynamics in the sequences. Nowadays, large efforts have been directed to the employment of deep neural networks. These architectures learn hierarchical layers of representations to perform pattern recognition and have demonstrated impressive results on many pattern recognition tasks (e.g., image classification \cite{Krizhevsky:2012} and face recognition \cite{Schroff:2015}). For instance, Recurrent Neural Networks (RNNs) with Long-Short Term Memory (LSTM) have been employed to model skeleton data for 3D action recognition~\cite{Veeriah:2015, Shahroudy:2016, Song:2017, Zhang:2017}. Although RNN approaches present excellent results in 3D action recognition task due to their power of modeling temporal sequences, such structures lack the ability to efficiently learn the spatial relations between the skeleton joints~\cite{Yang:2018}.

To take advantage of the spatial relations, a hierarchical structure was proposed by Du et al.~\cite{Du:2015}. The authors represent each skeleton sequence as 2D arrays, in which the temporal dynamics of the sequence is encoded as variations in columns and the spatial structure of each frame is represented as rows. Then, the representation is fed to a Convolutional Neural Network (CNN) which has the natural ability of learning structural information from 2D arrays. Such type of representations are very compact encoding the entire video sequence in one single image.

To further improve the representation of skeleton joints for 3D action recognition, in this paper we introduce a novel skeleton image representation to be used as input of CNNs named \metodo. The proposed approach encodes temporal dynamics by explicitly using motion information computing the magnitude and orientation values of the skeleton joints. To that end, different temporal scales are used to filter noisy motion values as well as aggregating more temporal dynamics to the representation making it being able to capture long-range joint interactions involved in actions. Moreover, the method takes advantage of a structural organization of joints that preserves spatial relations of more relevant joint pairs. To perform action classification, we train a tiny CNN architecture with only three convolutional layers and two fully-connected layers. Since the network is shallow and takes as input a compact representation for each video, it is extremely fast to train.

In the literature, many works employed or improved the skeleton image representation for 3D action recognition~\cite{Wang:2016, Liu:2017, Ke:2017, Li:2017, Wang:2018, Yang:2018, Li:2018, Choutas:2018}. However, none of the  methods model explicit motion information (i.e., magnitude and orientation) in multiple temporal scales, as the proposed approach does. Similar to our approach, the works of~\cite{Li:2017, Li:2018} were the only ones that tried to encode motion on skeleton images, however they employed a naive approach by computing difference of motion joints on consecutive frames.

According to the experimental results, our proposed skeleton image representation can handle skeleton based 3D action recognition very well. Moreover, \metodo~representation achieves the state-of-the-art performance on the large scale NTU RGB+D~120~\cite{Liu:2019} dataset when combined with a spatial structural joint representation.

The code of our \metodo~representation is publicly available to facilitate future research~\footnote{\url{https://github.com/carloscaetano/skeleton-images}}.

\section{Related Work}\label{related}

In this section, we present a literature review of works that are close to the idea proposed in our approach by employing different representations based on skeleton images.

As the forerunner of skeleton image representations, Du et al.~\cite{Du:2015} represent the skeleton sequences as a matrix. Each row of such matrix corresponds to a chain of concatenated skeleton joint coordinates from the frame $t$. Hence, each column of the matrix corresponds to the temporal evolution of the joint $j$. At this point, the matrix size is $J \times T \times 3$, where $J$ is the number of joints for each skeleton, $T$ is the total frame number of the video sequence and $3$ is the number coordinate axes ($x, y, z$). The values of this matrix are quantified into an image (i.e., linearly rescaled to a $[0, 255]$) and normalized to handle the variable-length problem. In this way, the temporal dynamics of the skeleton sequence is encoded as variations in rows and the spatial structure of each frame is represented as columns. Finally, the authors use their representation as input to a CNN model composed by four convolutional layers and three max-pooling layers. After the feature extraction, a feed-forward neural network with two fully-connected layers is employed for classification.

Wang et al.~\cite{Wang:2016, Wang:2018} present a skeleton representation to represent both spatial configuration and dynamics of joint trajectories into three texture images through color encoding, named Joint Trajectory Maps (JTMs). The authors apply rotations to the skeleton data to mimicking multi-views and also for data enlargement to overcome the drawback of CNNs usually being not view invariant. JTMs are generated by projecting the trajectories onto the three orthogonal planes. To encode motion direction in the JTM, they use a hue colormap function  to ``color'' the joint trajectories over the action period. They also encode the motion magnitude of joints into saturation and brightness claiming that changes in motion results in texture in the JMTs. Finally, the authors individually fine-tune three AlexNet~\cite{Krizhevsky:2012} CNNs (one for each JTM) to perform classification.

To overcome the problem of the sparse data generated by skeleton sequence video, Ke et al.~\cite{Ke:2017} represent the temporal dynamics of the skeleton sequence by generating four skeleton representation images. Their approach is closer to Du et al.~\cite{Du:2015} method, however they compute the relative positions of the joints to four reference joints by arranging them as a chain and concatenating the joints of each body part to the reference joints resulting onto four different skeleton representations. According to the authors, such structure incorporate different spatial relationships between the joints. Finally, the skeleton images are resized and each channel of the four representations is used as input to a VGG19~\cite{Simonyan:2015} pre-trained architecture for feature extraction.

To encode motion information on skeleton image representation, Li et al.~\cite{Li:2017} proposed the skeleton motion image. Their approach is created similar to Du et al.~\cite{Du:2015} skeleton image representation, however each matrix cell is composed by joint difference computation between two consecutive frames. To perform classification, the authors used Du et al.~\cite{Du:2015} approach and their proposed representation independently as input of a neural network with a two-stream paradigm. The CNN used was a small seven-layer network  consisting of three convolution layers and four fully-connected layers.

Yang et al.~\cite{Yang:2018} claim that the concatenation process of chaining all joints with a fixed order turn into lack of semantic meaning and leads to loss in skeleton structural information. To that end, Yang et al.~\cite{Yang:2018} proposed a representation named Tree Structure Skeleton Image (TSSI) to preserve spatial relations. Their method is created by traversing a skeleton tree with a depth-first order algorithm with the premise that the fewer edges there are, the more relevant the joint pair is. The generated representation are then quantified into an image and resized before being sent to a ResNet-50~\cite{He:2016} CNN architecture.

As it can be inferred from the reviewed methods, most of them are improved versions of Du et al.~\cite{Du:2015} skeleton image focusing on spatial structural of joint axes while the temporal dynamics of the sequence is encoded as variations in columns, or encode motion information in a naive manner (difference of motion joints on consecutive frames). Despite the aforementioned methods produce promising results, they do not explicit encode rich motion information. In view of that, to capture more motion information, our approach directly encodes it by using orientation and magnitude to provide information regarding the velocity of the movement in different temporal scales In view of that, our \metodo~approach differs from the literature methods by capturing the temporal dynamics explicitly provided by magnitude and orientation motion information.

\section{Proposed Approach}\label{approach}

\begin{figure*}[!t]
	\centering
	\includegraphics[width=1.0\textwidth]{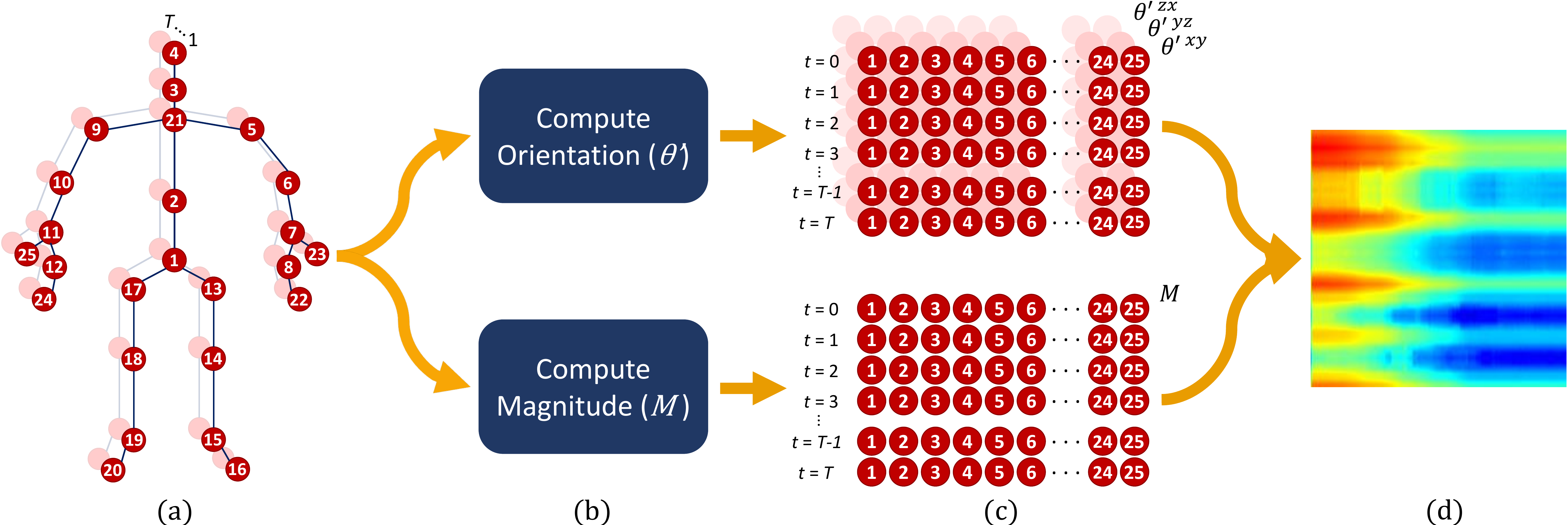}
	\caption{{\metodo~representation. (a) Skeleton data sequence of $T$ frames. (b) Computation of the magnitude and orientation from the joint movement. (c) $\theta^{'}$ and $M$ arrays: each row encodes the spatial information (relation between joint movements) while each column describes the temporal information for each joint movement. (d) Skeleton image after resizing and stacking of each axes.}}
	\label{img:sklimage}
\end{figure*}

In this section, we introduce our proposed skeleton image representation based on magnitude and orientation motion information, named \emph{\metodo}.

\subsection{\metodo}

As reviewed in Section~\ref{related}, the majority of works that encode skeleton data as image representations are based on spatial structure encoding of the skeleton joints. According to Li et al.~\cite{Li:2018}, temporal movements of joints can also be used as crucial cues for action recognition and although the temporal dynamics of the sequence can be implicitly learned by using a CNN, an explicit modeling can produce better recognition accuracies.

{Recently, a new temporal stream for the two-stream networks called Magnitude-Orientation Stream (MOS)~\cite{Caetano:2017} was developed. The method is based on non-linear transformations and claim that motion information on a video sequence can be described by the spatial relationship contained on the local neighborhood of magnitude and orientation extracted from the optical flow and has shown excellent results on the 2D action recognition problem. Motivated by such results, in this paper we propose a novel skeleton image representation (named \metodo), based on magnitude and orientation of the joints to explore the temporal dynamics. Our approach expresses the displacement information by using orientation encoding (direction of joints) and magnitude to provide information regarding the velocity of the movement.} Furthermore, due to the successful results achieved by the skeleton image representations, our approach follows the same fundamentals by representing the skeleton sequences as a matrix. First, we apply the depth-first tree traversal order~\cite{Yang:2018} to the skeleton joints to generate a pre-defined chain order $C$ that best preserves the spatial relations between joints in original skeleton structures\footnote{Chain $C$ considering 25 Kinect joints: [2, 21, 3, 4, 3, 21, 5, 6, 7, 8, 22, 23, 22, 8, 7, 6, 5, 21, 9, 10, 11, 12, 24, 25, 24, 12, 11, 10, 9, 21, 2, 1, 13, 14, 15, 16, 15, 14, 13, 1, 17, 18, 19, 20, 19, 18, 17, 1, 2], as defined in~\cite{Yang:2018}.}. Afterwards, we compute matrix $S$ that corresponds to a chain of concatenated skeleton joint coordinates from the frame $t$. In view of that, each column of the matrix corresponds to the temporal evolution of the arranged chain joint $c$. At this point, the size of matrix $S$ is $C \times T \times 3$, where $C$ is the number of joints of the chain, $T$ is the total frame number of the video sequence and $3$ is the number joint coordinate axes ($x, y, z$). Then, we create the motion structure $\mathcal D$ as
\begin{equation}\label{eq:dif}
	\mathcal D_{c,t} = S_{c,t+d} - S_{c},
\end{equation}
\noindent {where each matrix cell is composed by the temporal difference computation of each joint between two frames of $d$ distance, resulting in a $C \times T-d \times 3$ matrix.} 

{By using the proposed motion structure $\mathcal D$, we build two different representations: one based on the magnitudes of joint motions and another one based the orientations of the joint motion.} We compute both representations using
\begin{equation}\label{eq:mag}
	M_{c,t} = \sqrt{ (\mathcal D^{x}_{c,t})^{2} + (\mathcal D^{y}_{c,t})^{2} + (\mathcal D^{z}_{c,t})^{2}}
\end{equation}
and 
\begin{equation}\label{eq:ori_}
	\theta_{c,t} = \stack(\theta^{xy}_{c,t}, \theta^{yz}_{c,t}, \theta^{zx}_{c,t}) \nonumber
\end{equation}

\begin{equation}\label{eq:ori}
	\begin{array}{rl}
		\theta^{xy}_{c,t} = \tan^{-1} \left( \frac{\mathcal D^{y}_{c,t}}{\mathcal D^{x}_{c,t}} \right),\\
		&\mbox{  } \\
		\theta^{yz}_{c,t} = \tan^{-1} \left( \frac{\mathcal D^{z}_{c,t}}{\mathcal D^{y}_{c,t}} \right),\\
		&\mbox{  } \\
		\theta^{zx}_{c,t} = \tan^{-1} \left( \frac{\mathcal D^{x}_{c,t}}{\mathcal D^{z}_{c,t}} \right),
	\end{array}
\end{equation}

\noindent where $M$ is the magnitude skeleton representation of size $J \times T-d \times 1$ and $\theta$ is the orientation skeleton representation of size $J \times T-d \times 3$ (composed by 3 stacked channels).

Since the orientation values are estimated for every joint, it might generate noisy values for joints without any movement. Therefore, we perform a filtering on $\theta$ based on the values of $M$ as 
\begin{equation}\label{eq:orifilter}
	\theta^{'}_{c,t} = \left\{
	\begin{array}{rl}
		0, &\mbox{ if $M_{c,t} < m$} \\
		\theta_{c,t}, &\mbox{ otherwise }
	\end{array} \right.,
\end{equation}
\noindent where $m$ is a magnitude threshold value.

Finally, the generated matrices are normalized into [0, 1] and empirically resized into a fixed size of $C \times 100$, since number of frames may vary depending on the skeleton sequence of each video. Figure~\ref{img:sklimage} gives an overview of our method for building the \metodo~representation.

\subsubsection{Temporal Scale Aggregation (TSA)}

{Skeleton image representations in the literature basically encodes joint coordinates as channels. In view of that, it may cause a problem that the co-occurrence features are aggregated locally, being not able to capture long-range joint interactions involved in actions~\cite{Li:2018}.} Moreover, one drawback of encoding motion values of joints is the noisy values that can be introduced to the representation due to small distance $d$ between two frames. For instance, if the computation is performed considering two consecutive frames, it could add to the representation unnecessary motion of joints that are irrelevant to predict a specific action (e.g., motion of the head joint on a handshake action).

{To overcome the aforementioned problems, we also propose a variation of our \metodo~representation by pre-computing the motion structure $\mathcal D$ considering different $d$ distances.} For each of the motion structures $\mathcal D$, we compute its respective magnitude skeleton representation $M$ and then stack them all into one single representation. The same is applied to compute the orientation skeleton representation $\theta$, however a weighting scheme is applied during the filtering process explained before, as
\begin{equation}\label{eq:weightingorifilter}
	\theta^{'}_{c,t} = \left\{
	\begin{array}{rl}
		0, &\mbox{ if $M_{c,t} < m \times d$} \\
		\theta_{c,t}, &\mbox{ otherwise }
	\end{array} \right..
\end{equation}
{Such technique adds more temporal dynamics to the representation by explicitly showing temporal scales to the network. In this way, the network can learn which movements are really important for the action learning and also being able to capture long-range joint interactions.}

\section{Experimental Results}\label{experiments}

\begin{figure*}[!t]
	\centering
	\includegraphics[width=1.0\textwidth]{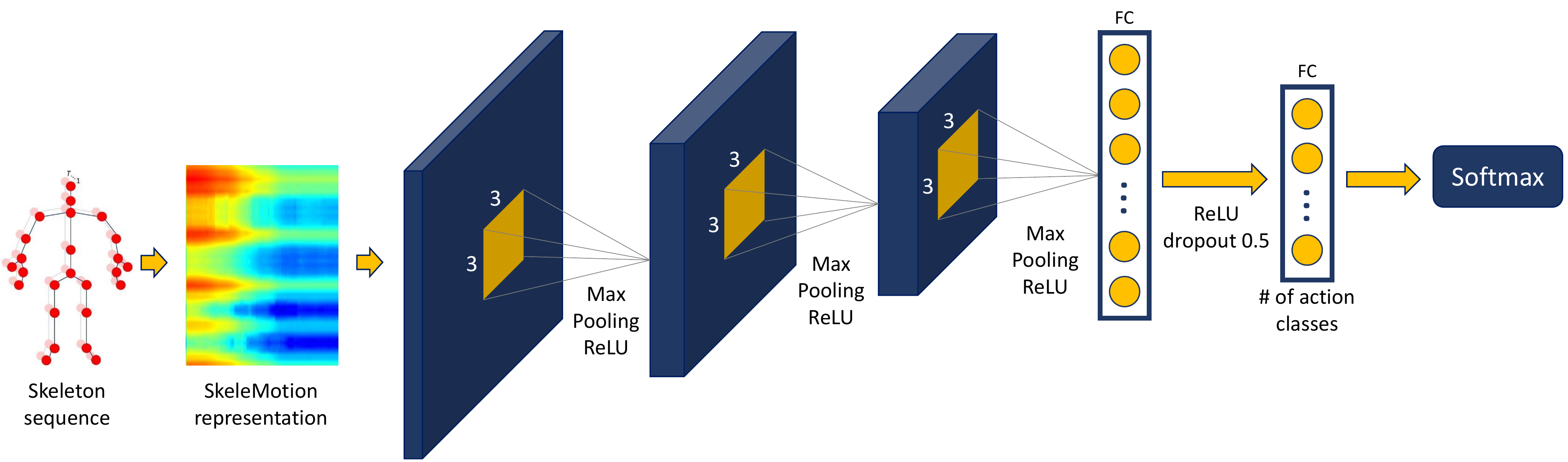}
	\caption{Network architecture employed for 3D action recognition.}
	\label{img:architecture}
\end{figure*}

In this section we present the experimental results obtained with the proposed skeleton image representation for the 3D action recognition problem. We compare it to other skeleton representations in the literature. Besides the classical skeleton image representation of Du et al.~\cite{Du:2015}, we compare with other representations used by state-of-the-art approaches~\cite{Wang:2016, Ke:2017, Li:2017, Li:2018, Yang:2018} as baselines on {NTU~RGB+D~60}~\cite{Shahroudy:2016}. We also compare our approach to sate-of-the-art methods on the {NTU~RGB+D~120}~\cite{Liu:2019}. To isolate only the contribution brought by \metodo~to the action recognition problem, all other representations were tested on the same datasets with the same split of training and testing data and using the same CNN. 

\subsection{Datasets}

The \textbf{NTU RGB+D 60}~\cite{Shahroudy:2016} is publicly available 3D action recognition dataset. It consists of 56,880 videos from 60 action categories which are performed by 40 distinct subjects. The videos were collected by three Microsoft Kinect sensors. The dataset provides four different data information: (i) RGB frames; (ii) depth maps; (iii) 395 infrared sequences; and (iv) skeleton joints. There are two different evaluation protocols: cross-subject, which split the 40 subjects into training and testing; and cross-view, which uses samples from one camera for testing and the other two for training. The performance is evaluated by computing the average recognition across all classes.

The \textbf{NTU RGB+D 120}~\cite{Liu:2019} is a large-scale 3D action recognition dataset captured under various environmental conditions. It consists of 114,480 RGB+D video samples captured using the Microsoft Kinect sensor. As in NTU RGB+D 60~\cite{Shahroudy:2016}, the dataset provides RGB frames, depth maps, infrared sequences and skeleton joints. It is composed by 120 action categories performed by 106 distinct subjects in a wide range of age distribution. There are two different evaluation protocols: cross-subject, which split the 106 subjects into training and testing; and cross-setup, which divides samples with even setup IDs for training (16 setups) and odd setup IDs for testing (16 setups). The performance is evaluated by computing the average recognition across all classes.

\subsection{Implementation Details}

To isolate only the contribution brought by the proposed representation to the action recognition problem, all compared skeleton image representations were implemented and tested on the same datasets and using the same network architecture. In view of that, we applied the same split of training and testing data, and employ the evaluation protocols and metrics proposed by the creators of the datasets.

The network architecture employed is a modified version of the CNN proposed by Li et al.~\cite{Li:2017}. They designed a small convolutional neural network which consists of three convolution layers and four fully-connected (FC) layers. However, here we modified it to a tiny version, employing the convolutional layers and only two FC layers. All convolutions have a kernel size of $3 \times 3$, the first and second convolutional layers with a stride of 1 and the third one with a stride of 2. Max pooling and ReLU neuron are adopted and the dropout regularization ratio is set to 0.5. The learning rate is set to 0.001 and batch size is set to 1000. The training is stopped after 200 epochs. The loss function employed was the categorical cross-entropy. We opted for using such architecture since it demonstrated good performance and, according to the authors, it can be easily trained from scratch without any pre-training and is superior on its compact model size and fast inference speed as well. Figure~\ref{img:architecture} presents an overview of the employed architecture.

To cope with actions involving multi-person interaction (e.g., shaking hands), we apply a common choice in the literature which is to stack skeleton image representations of different people as the network input.

To obtain the orientation skeleton image representation $\theta^{'}$ we empirically set the parameter $m = 0.004$, as described in Section~\ref{approach}.

\subsection{Evaluation}

In this section, we present experiments for parameters optimization and report a comparison of our proposed skeleton representation. We used a subset of NTU RGB+D~60~\cite{Shahroudy:2016} training set (considering cross-view protocol) to perform parameter setting and then used such parameter on the remaining experiments. We focused on the optimization of the number of temporal scales used on temporal scale aggregation (TSA).

\begin{table}[th!]
	\centering
	\begin{small}
		\caption{Action recognition accuracy (\%) results on a subset of NTU RGB+D~60~\cite{Shahroudy:2016} dataset by applying temporal scale aggregation (TSA) on our~\metodo~representation.}
		\begin{tabular}{clcc}
			\toprule
			& \multicolumn{1}{l}{\textbf{Temporal}} & \multicolumn{1}{c}{\textbf{Magnitude}} & \multicolumn{1}{c}{\textbf{Orientation}}\\
			& \multicolumn{1}{l}{\textbf{distances}} & \textbf{Acc. (\%)} & \textbf{Acc. (\%)}\\
			\toprule
			& 1, 5 & 64.9 & 62.4 \\
			\multirow{1}{*}{\textbf{Two Temporal}} & 1, 10 & 67.4 & 62.9\\
			\multirow{1}{*}{\textbf{Scales}} & 1, 15 & 66.0 & 64.1\\
			& 1, 20 & 66.1 & 63.5\\
			\midrule
			\multirow{2}{*}{\textbf{Three Temporal}} & 1, 5, 10 & 68.6 & 64.6\\
			\multirow{2}{*}{\textbf{Scales}} & 1, 10, 20 & 69.0 & \textbf{65.4}\\
			& 5, 10, 15 & \textbf{70.1} & 65.1\\ 
			\midrule
			\multirow{1}{*}{\textbf{Four Temporal}} & 1, 5, 10, 15 & 69.6 & 65.2 \\
			\multirow{1}{*}{\textbf{Scales}} & 5, 10, 15, 20 & 67.9 & 64.4\\
			
			\bottomrule
		\end{tabular}
		\label{tab:mag-temporal}
	\end{small}
\end{table}

\begin{figure*}[!htb]
	\centering
	\includegraphics[width=1.0\textwidth]{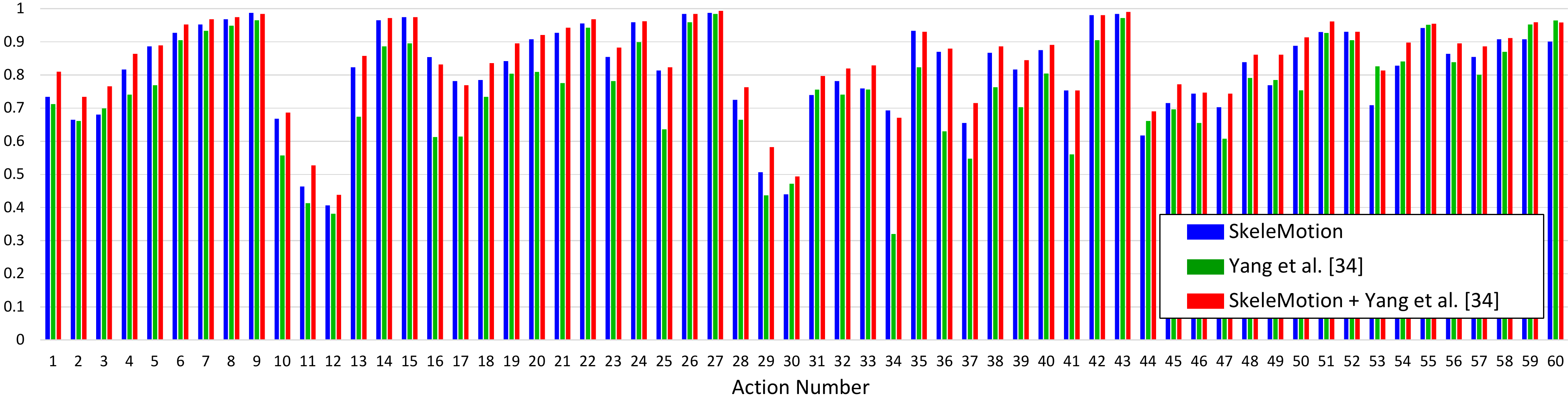}
	\caption{The complementary between \metodo~(Magnitude-Orientation TSA) and Yang et al.~\cite{Yang:2018} TSSI representation on NTU RGB+D~60~\cite{Shahroudy:2016} dataset for cross-view protocol. Best viewed in color.}
	\label{img:complementary}
\end{figure*}

To set the number of temporal scales of our \metodo~approach, we empirically varied it from two to four temporal scales considering 20 frames in total. Table~\ref{tab:mag-temporal} shows the results obtained by such variation. We can see that the best result is obtained by using three temporal scales for both magnitude (5, 10, 15) and orientation (1, 10, 20). Moreover, we noticed that the performance tends to saturate or drop when considering four temporal scales.

Table~\ref{tab:baseline-comparison} presents a comparison of our approach with skeleton image representations of the literature. For the methods that have more than one ``image'' per representation (\cite{Wang:2016} and \cite{Ke:2017}), we stacked them to be used as input to the network. The same was performed for our \metodo~approach considering magnitude and orientation. Regarding the cross-subject protocol, the best result was obtained by Reference Joints representation  from Ke~et~al.,~\cite{Ke:2017} achieving 70.8\% of accuracy while our best result (\metodo~Magnitude (TSA)) achieves a competitive accuracy of 69.6\%. It is worth noting that there is a considerable improvement of 12.8 (p.p.) obtained by \metodo~Magnitude (TSA) when compared to Li~et~al.,~\cite{Li:2017} baseline, which also explicitly encode motion information. On the other side, the best result on cross-view protocol was obtained by our \metodo~Magnitude (TSA) approach achieving 80.1\% of accuracy. There is an improvement of 4.5 percentage points (p.p.) when compared to the Tree Structure Skeleton Image (TSSI) from Yang~et~al.,~\cite{Yang:2018}, which was the best baseline result. Again, there is a considerable improvement of 18.8 (p.p.) when compared to Li~et~al.,~\cite{Li:2017} baseline.

\begin{table}[th!]
	\centering
	\begin{small}
		\caption{Action recognition accuracy (\%) results on NTU RGB+D~60~\cite{Shahroudy:2016} dataset. Results for the baselines were obtained running each method implementation.}
		\begin{tabular}{clcc}
			\toprule
			& & \multicolumn{1}{c}{\textbf{Cross-}} & \multicolumn{1}{c}{\textbf{Cross-}} \\
			& & \multicolumn{1}{c}{\textbf{subject}} & \multicolumn{1}{c}{\textbf{view}} \\
			& \textbf{Approach} & \multicolumn{1}{c}{\textbf{Acc. (\%)}} & \multicolumn{1}{c}{\textbf{Acc. (\%)}} \\
			\toprule
			& Du et al.~\cite{Du:2015} & 68.7 & 73.0 \\
			& Wang et al.~\cite{Wang:2016} & 39.1 & 35.9 \\ 
			\multirow{1}{*}{\textbf{Baselines}} & Ke et al.~\cite{Ke:2017} & \textbf{70.8} & 75.5 \\
			& Li et al.~\cite{Li:2018} & 56.8 & 61.3 \\
			& Yang et al.~\cite{Yang:2018} & 69.5 & \textbf{75.6} \\
			\midrule
			& Orientation & 60.6 & 65.6 \\
			\multirow{1}{*}{\textbf{\metodo}} & Magnitude & 58.4 & 64.2 \\
			\multirow{1}{*}{\textbf{results}} & Orientation (TSA) & 65.3 & 73.2 \\ 
			& Magnitude (TSA) & \textbf{69.6} & \textbf{80.1} \\ 
			\bottomrule
		\end{tabular}
		\label{tab:baseline-comparison}
	\end{small}
\end{table}

To exploit a possible complementarity of the temporal (our \metodo) and spatial (Yang et al.~\cite{Yang:2018}) skeleton image representations, we combined the different approaches by employing early and late fusion techniques. For the early fusion, we simply stacked the representations to be used as input to the network. On the other hand, the late fusion technique applied was a non-weighted linear combination of the prediction scores of each method. According to the results showed in Table~\ref{tab:fusions}, any type of combination performed with our \metodo~provides better results than their solo versions. 
Regarding cross-subject protocol, our best results achieves 73.5\% of accuracy with early fusion technique against 76.5\% of the late fusion approach. Furthermore, on cross-view protocol, our best results achieves 82.4\% of accuracy with early fusion technique against 84.7\% of the late fusion approach. Detailed improvements are shown in Figure~\ref{img:complementary}.

\begin{table*}[th!]
	\centering
	\begin{small}
		\caption{Comparison between late and early fusion techniques on NTU RGB+D~60~\cite{Shahroudy:2016} dataset.}
		\begin{tabular}{clcc}
			\toprule
			& & \multicolumn{1}{c}{\textbf{Cross-subject}} & \multicolumn{1}{c}{\textbf{Cross-view}} \\
			& \textbf{Approach} & \multicolumn{1}{c}{\textbf{Acc. (\%)}} & \multicolumn{1}{c}{\textbf{Acc. (\%)}} \\
			\toprule
			& Magnitude-Orientation & \multicolumn{1}{c}{ 65.6 } & \multicolumn{1}{c}{ 71.1 } \\
			& Magnitude-Orientation $+$ Yang et al.\cite{Yang:2018} & \multicolumn{1}{c}{ 70.1 } & \multicolumn{1}{c}{ 78.8 } \\
			\multirow{1}{*}{\textbf{Early}} & Magnitude-Orientation (TSA) & \multicolumn{1}{c}{ 70.5 } & \multicolumn{1}{c}{ 78.7 } \\
			\multirow{1}{*}{\textbf{Fusion}} & Magnitude (TSA) $+$ Yang et al.\cite{Yang:2018} & \multicolumn{1}{c}{ 71.7 } & \multicolumn{1}{c}{ \textbf{82.4} } \\
			& Orientation (TSA) $+$ Yang et al.\cite{Yang:2018} & \multicolumn{1}{c}{ 69.6 } & \multicolumn{1}{c}{ 78.9 } \\
			& Magnitude-Orientation (TSA) $+$ Yang et al.\cite{Yang:2018} & \multicolumn{1}{c}{ \textbf{73.5} } & \multicolumn{1}{c}{ 82.1 } \\
			\midrule
			& Magnitude-Orientation & \multicolumn{1}{c}{ 65.6 } & \multicolumn{1}{c}{ 71.1 } \\
			& Magnitude-Orientation $+$ Yang et al.\cite{Yang:2018} & \multicolumn{1}{c}{ 73.2 } & \multicolumn{1}{c}{ 79.5 } \\
			\multirow{1}{*}{\textbf{Late}} & Magnitude-Orientation (TSA) & \multicolumn{1}{c}{ 72.2 } & \multicolumn{1}{c}{ 81.7 } \\
			\multirow{1}{*}{\textbf{Fusion}} & Magnitude (TSA) $+$ Yang et al.\cite{Yang:2018} & \multicolumn{1}{c}{ 75.4 } & \multicolumn{1}{c}{ 83.2 } \\ 
			& Orientation (TSA) $+$ Yang et al.\cite{Yang:2018} & \multicolumn{1}{c}{ 73.6 } & \multicolumn{1}{c}{ 80.6 } \\
			& Magnitude-Orientation (TSA) $+$ Yang et al.\cite{Yang:2018} & \multicolumn{1}{c}{ \textbf{76.5} } & \multicolumn{1}{c}{ \textbf{84.7} } \\			
			\bottomrule
		\end{tabular}
		\label{tab:fusions}
	\end{small}
\end{table*}

\begin{table*}[th!]
	\centering
	\begin{small}
		\caption{Action recognition accuracy (\%) results on NTU RGB+D~120~\cite{Liu:2019} dataset. Results for literature methods were obtained from~\cite{Liu:2019}.}
		\begin{tabular}{clcc}
			\toprule
			& & \multicolumn{1}{c}{\textbf{Cross-subject}} & \multicolumn{1}{c}{\textbf{Cross-setup}} \\
			& \textbf{Approach} & \multicolumn{1}{c}{\textbf{Acc. (\%)}} & \multicolumn{1}{c}{\textbf{Acc. (\%)}} \\
			\toprule
			
			& Part-Aware LSTM~\cite{Shahroudy:2016} & 25.5 & 26.3 \\
			& Soft RNN~\cite{Hu:2018} & 36.3 & 44.9 \\
			& Dynamic Skeleton~\cite{Hu:2017} & 50.8 & 54.7 \\
			& Spatio-Temporal LSTM~\cite{Liu:2016} & 55.7 & 57.9 \\
			& Internal Feature Fusion~\cite{Liu:2018} & 58.2 & 60.9 \\
			\multirow{1}{*}{\textbf{Literature}} & GCA-LSTM~\cite{Liu:2017b} & 58.3 & 59.2 \\
			\multirow{1}{*}{\textbf{results}} & Multi-Task Learning Network~\cite{Ke:2017} & 58.4 & 57.9 \\
			& FSNet~\cite{Liu:2019b} & 59.9 & 62.4 \\
			& Skeleton Visualization (Single Stream)~\cite{Liu:2017c} & 60.3 & 63.2 \\
			& Two-Stream Attention LSTM~\cite{Liu:2018b} & 61.2 & 63.3 \\
			& Multi-Task CNN with RotClips~\cite{Ke:2018} & 62.2 & 61.8 \\
			& Body Pose Evolution Map~\cite{Liu:2018c} & \textbf{64.6} & \textbf{66.9} \\
			
			\midrule
			
			& Orientation (TSA) & 52.2 & 54.1 \\ 
			\multirow{1}{*}{\textbf{\metodo}} & Magnitude (TSA) & 57.6 & 60.4 \\
			\multirow{1}{*}{\textbf{results}} & Magnitude-Orientation (TSA) & 62.9 & 63.0 \\
			& Magnitude-Orientation (TSA) $+$ Yang et al.\cite{Yang:2018} & \textbf{ 67.7 } & \textbf{ 66.9 } \\ 
			
			\bottomrule
		\end{tabular}
		\label{tab:NTU120-comparison}
	\end{small}
\end{table*}

Finally, Table~\ref{tab:NTU120-comparison} presents the experiments of our proposed skeleton image representation on the recent available NTU RGB+D~120~\cite{Liu:2019} dataset. Due to the results obtained on Table~\ref{tab:fusions}, here we employed the late fusion scheme for methods combination.

We obtained good results with our \metodo~representation outperforming many skeleton based methods~\cite{Shahroudy:2016, Hu:2018, Hu:2017, Liu:2016, Liu:2018, Liu:2017b, Ke:2017, Liu:2019b, Liu:2017c, Liu:2018b, Ke:2018, Liu:2018c}. When combining our representation with Yang et al.~\cite{Yang:2018} we achieve state-of-the-art results, outperforming the best reported method (Body Pose Evolution Map~\cite{Liu:2018c}) by up to 3.1 p.p. on cross-subject protocol and achieve competitive results on cross-setup protocol.

In comparison with LSTM approaches, we outperform the best reported method (Two-Stream Attention LSTM) by 1.7 p.p. using our skeleton image representation and 6.5 p.p. when combining it with Yang et al.~\cite{Yang:2018} method on cross-subject protocol. Regarding the cross-setup protocol, we outperform them by 3.6 p.p. using our skeleton image representation fused with Yang et al.~\cite{Yang:2018}. This indicates that our skeleton image representation approach used as input for CNNs leads to a better learning of temporal dynamics than the approaches that employs LSTM.

\section{Conclusions and Future Works}\label{conclusions}

In this work, we proposed a novel skeleton image representation to be used as input of CNNs, named \metodo. The method is based on temporal dynamics encoding by explicitly using motion information (magnitude and orientation) of the skeleton joints. We further propose a variation of the magnitude skeleton representation considering different temporal scales in order to filter noisy motion values as well as aggregating more temporal dynamics to the representation. Experimental results on two publicly available datasets demonstrated the excellent performance of the proposed approach. Another interesting finding is that the combination of our representation with methods of the literature improves the 3D action recognition outperforming the state-of-the-art on NTU RGB+D~120 dataset.

Directions to future works include the evaluation of \metodo~with other distinct architectures. Moreover, we intend to evaluate its behavior on 2D action datasets with skeletons estimated by methods of the literature.

\section*{Acknowledgments}

The authors would like to thank the National Council for Scientific and Technological Development -- CNPq (Grants~311053/2016-5, 204952/2017-4 and~438629/2018-3), the Minas Gerais Research Foundation -- FAPEMIG (Grants~APQ-00567-14 and~PPM-00540-17) and the Coordination for the Improvement of Higher Education Personnel -- CAPES (DeepEyes Project).

\balance
{\small
\bibliographystyle{ieee}
\bibliography{bibliography}
}

\end{document}